# Frame-Recurrent Video Super-Resolution

Mehdi S. M. Sajjadi[1,2][*]  Raviteja Vemulapalli[2]  Matthew Brown[2]
msajjadi@tue.mpg.de  ravitejavemu@google.com  mtbr@google.com

[1]Max Planck Institute for Intelligent Systems  [2]Google

## Abstract

*Recent advances in video super-resolution have shown that convolutional neural networks combined with motion compensation are able to merge information from multiple low-resolution (LR) frames to generate high-quality images. Current state-of-the-art methods process a batch of LR frames to generate a single high-resolution (HR) frame and run this scheme in a sliding window fashion over the entire video, effectively treating the problem as a large number of separate multi-frame super-resolution tasks. This approach has two main weaknesses: 1) Each input frame is processed and warped multiple times, increasing the computational cost, and 2) each output frame is estimated independently conditioned on the input frames, limiting the system's ability to produce temporally consistent results.*

*In this work, we propose an end-to-end trainable frame-recurrent video super-resolution framework that uses the previously inferred HR estimate to super-resolve the subsequent frame. This naturally encourages temporally consistent results and reduces the computational cost by warping only one image in each step. Furthermore, due to its recurrent nature, the proposed method has the ability to assimilate a large number of previous frames without increased computational demands. Extensive evaluations and comparisons with previous methods validate the strengths of our approach and demonstrate that the proposed framework is able to significantly outperform the current state of the art.*

## 1. Introduction

Super-resolution is a classic problem in image processing that addresses the question of how to reconstruct a high-resolution (HR) image from its downscaled low-resolution (LR) version. With the rise of deep learning, super-resolution has received significant attention from the research community over the past few years [3, 5, 20, 21, 26, 28, 35, 36, 39]. While high-frequency details need to be reconstructed exclusively from spatial statistics in the case of single image super-resolution, temporal relationships in the input can be ex-ploited to improve reconstruction for video super-resolution. It is therefore imperative to combine the information from as many LR frames as possible to reach the best video super-resolution results.

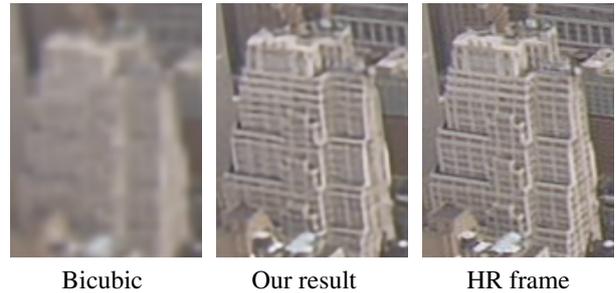

Figure 1: Side-by-side comparison of bicubic interpolation, our FRVSR result, and HR ground truth for 4x upsampling.

The latest state-of-the-art video super-resolution methods approach the problem by combining a batch of LR frames to estimate a single HR frame, effectively dividing the task of video super-resolution into a large number of separate multi-frame super-resolution subtasks [3, 28, 29, 39]. However, this approach is computationally expensive since each input frame needs to be processed several times. Furthermore, generating each output frame separately reduces the system's ability to produce temporally consistent frames, resulting in unpleasing flickering artifacts.

In this work, we propose an end-to-end trainable frame-recurrent video super-resolution (FRVSR) framework to address the above issues. Instead of estimating each video frame separately, we use a recurrent approach that passes the previously estimated HR frame as an input for the following iteration. Using this recurrent architecture has several benefits. Each input frame needs to be processed only once, reducing the computational cost. Furthermore, information from past frames can be propagated to later frames via the HR estimate that is recurrently passed through time. Passing the previous HR estimate directly to the next step helps the model to recreate fine details and produce temporally consistent videos.

[*]This work was done when Mehdi S. M. Sajjadi was interning at Google.

To analyze the performance of the proposed framework, we compare it with strong single image and video super-resolution baselines using identical neural networks as building blocks. Our extensive set of experiments provides insights into how the performance of FRVSR varies with the number of recurrent steps used during training, the size of the network, and the amount of noise, aliasing or compression artifacts present in the LR input. The proposed approach clearly outperforms the baselines under various settings both in terms of quality and efficiency. Finally, we also compare FRVSR with several existing video super-resolution approaches and show that it significantly outperforms the current state of the art on a standard benchmark dataset.

### 1.1. Our contributions

• We propose a recurrent framework that uses the HR estimate of the previous frame for generating the subsequent frame, leading to an efficient model that produces temporally consistent results.

• Unlike existing approaches, the proposed framework can propagate information over a large temporal range without increasing computations.

• Our system is end-to-end trainable and does not require any pre-training stages.

• We perform an extensive set of experiments to analyze the proposed framework and relevant baselines under various different settings.

• We show that the proposed framework significantly outperforms the current state of the art in video super-resolution both qualitatively and quantitatively.

## 2. Video super-resolution

Let $I_t^{\text{LR}} \in [0, 1]^{H \times W \times C}$ denote the $t$-th LR video frame obtained by downsampling the original HR video frame $I_t^{\text{HR}} \in [0, 1]^{sH \times sW \times C}$ by scale factor $s$. Given a set of consecutive LR video frames, the goal of video super-resolution is to generate HR estimates $I_t^{\text{est}}$ that approximate the original HR frames $I_t^{\text{HR}}$ under some metric.

### 2.1. Related work

Super-resolution is a classic ill-posed inverse problem with approaches ranging from simple interpolation methods such as Bilinear, Bicubic and Lanczos [9] to example-based super-resolution [12, 13, 40, 42], dictionary learning [32, 43], and self-similarity approaches [16, 41]. We refer the reader to Milanfar [30] and Nasrollahi and Moeslund [31] for extensive overviews of prior art up to recent years.

The recent progress in deep learning, especially in convolutional neural networks, has shaken up the field of super-resolution. After Dong et al. [5] reached state-of-the-art results with shallow convolutional neural networks, many others followed up with deeper network architectures, advancing the field tremendously [6, 21, 22, 25, 36, 37]. Parallel efforts have studied alternative loss functions for more visually pleasing reconstructions [26, 35]. Agustsson and Timofte [1] provide a recent survey on the current state of the art in single image super-resolution.

Video and multi-frame super-resolution approaches combine information from multiple LR frames to reconstruct details that are missing in individual frames which can lead to higher quality results. Classical video and multi-frame super-resolution methods are generally formulated as optimization problems that are computationally very expensive to solve [2, 11, 27, 38].

Most of the existing deep learning-based video super-resolution methods divide the task of video super-resolution into multiple separate sub-tasks, each of which generates a single HR output frame from multiple LR input frames. Kappeler et al. [20] warp video frames $I_{t-1}^{\text{LR}}$ and $I_{t+1}^{\text{LR}}$ onto the frame $I_t^{\text{LR}}$ using the optical flow method of Drulea and Nedevschi [8], concatenate the three frames and pass them through a convolutional neural network that produces the output frame $I_t^{\text{est}}$. Caballero et al. [3] follow the same approach but replace the optical flow model with a trainable motion compensation network. Makansi et al. [29] follow an approach similar to [3] but combine warping and mapping to HR space into a single step.

Tao et al. [39] rely on a batch of up to 7 input LR frames to estimate a single HR frame. After computing the motion from neighboring input frames to $I_t^{\text{LR}}$, they map the frames onto high-resolution grids. In a final step, they run an encoder-decoder style network with a Conv-LSTM in the core yielding $I_t^{\text{est}}$. Liu et al. [28] process up to 5 LR frames using different numbers of input frames ($I_t^{\text{LR}}$), ($I_{t-1}^{\text{LR}}, I_t^{\text{LR}}, I_{t+1}^{\text{LR}}$), and ($I_{t-2}^{\text{LR}}, \ldots, I_{t+2}^{\text{LR}}$) simultaneously to produce separate HR estimates that are aggregated in a final step with dynamic weights to produce a single output $I_t^{\text{est}}$.

While a number of the above mentioned methods are end-to-end trainable, the authors often note that they first pre-train each component before fine-tuning the system as a whole in a final step [3, 28, 39].

Huang et al. [17] use a bidirectional recurrent architecture for video super-resolution with shallow networks but do not use any explicit motion compensation in their model. Recurrent architectures have also been used for other tasks such as video deblurring [23] and stylization [4, 15]. While Kim et al. [23] and Chen et al. [4] pass on a feature representation to the next step, Gupta et al. [15] pass the previous output frame to the next step to produce temporally consistent stylized videos in concurrent work. A recurrent approach for video super-resolution was proposed by Farsiu et al. [10] more than a decade ago with motivations similar to ours. However, this approach uses an approximation of the Kalman filter for frame estimation and is constrained to translational motion.

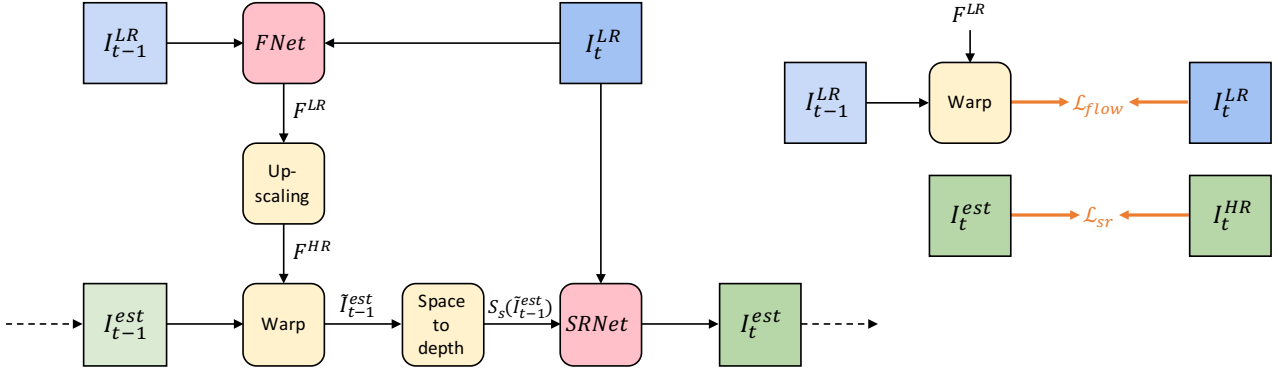

Figure 2: Overview of the proposed FRVSR framework (left) and the loss functions used for training (right). After computing the flow $F^{\text{LR}}$ in LR space using FNet, we upsample it to $F^{\text{HR}}$. We then use $F^{\text{HR}}$ to warp the HR-estimate of the previous frame $I^{\text{est}}_{t-1}$ onto the current frame. Finally, we map the warped previous output $\tilde{I}^{\text{est}}_{t-1}$ to LR-space using the space-to-depth transformation and feed it to the super-resolution network SRNet along with the current input frame $I^{\text{LR}}_t$. For training the networks (shown in red), we apply a loss on $I^{\text{est}}_t$ as well as an additional loss on the warped previous LR frame to aid FNet.

## 3. Method

After presenting an overview of the FRVSR framework in Sec. 3.1 and defining the loss functions used for training in Sec. 3.2, we justify our design choices in Sec. 3.3 and give details on the implementation and training procedure in Sec. 3.4 and 3.5, respectively.

### 3.1. FRVSR Framework

The proposed framework is illustrated in Fig. 2. Trainable components (shown in red) include the optical flow estimation network FNet and the super-resolution network SRNet. To produce the HR estimate $I^{\text{est}}_t$, our model makes use of the current LR input frame $I^{\text{LR}}_t$, the previous LR input frame $I^{\text{LR}}_{t-1}$, and the previous HR estimate $I^{\text{est}}_{t-1}$.

**1. Flow estimation:** As a first step, FNet estimates the flow between the low-resolution inputs $I^{\text{LR}}_{t-1}$ and $I^{\text{LR}}_t$ yielding the normalized low-resolution flow map

$$F^{\text{LR}} = \text{FNet}(I^{\text{LR}}_{t-1}, I^{\text{LR}}_t) \in [-1,1]^{H \times W \times 2} \quad (1)$$

that assigns a position in $I^{\text{LR}}_{t-1}$ to each pixel location in $I^{\text{LR}}_t$.

**2. Upscaling flow:** Treating the flow map $F^{\text{LR}}$ as an image, we upscale it using bilinear interpolation with scaling factor $s$ which results in an HR flow-map

$$F^{\text{HR}} = \text{UP}(F^{\text{LR}}) \in [-1,1]^{sH \times sW \times 2}. \quad (2)$$

**3. Warping previous output:** We use the high-resolution flow map $F^{\text{HR}}$ to warp the previously estimated image $I^{\text{est}}_{t-1}$ according to the optical flow from the previous frame onto the current frame.

$$\tilde{I}^{\text{est}}_{t-1} = \text{WP}(I^{\text{est}}_{t-1}, F^{\text{HR}}) \quad (3)$$

We implemented warping as a differentiable function using bilinear interpolation similar to Jaderberg *et al.* [19].

**4. Mapping to LR space:** We map the warped previous output $\tilde{I}^{\text{est}}_{t-1}$ to LR space using the space-to-depth transformation

$$S_s : [0,1]^{sH \times sW \times C} \to [0,1]^{H \times W \times s^2 C} \quad (4)$$

which extracts shifted low-resolution grids from the image and places them into the channel dimension, see Fig. 3 for an illustration. The operator can be formally described as

$$S_s(I)_{i,j,k} = I_{si+k\%s,\ sj+(k/s)\%s,\ k/s^2} \quad (5)$$

with zero-based indexing, modulus $\%$ and integer division $/$.

**5. Super-Resolution:** In the final step, we concatenate the LR mapping of the warped previous output $\tilde{I}^{\text{est}}_{t-1}$ with the current low-resolution input frame $I^{\text{LR}}_t$ in the channel dimension, and feed the result $I^{\text{LR}}_t \oplus S_s(\tilde{I}^{\text{est}}_{t-1})$ to the super-resolution network SRNet.

**Summary:** The final estimate $I^{\text{est}}_t$ of the framework is the output of the super-resolution network SRNet:

$$\text{SRNet}(I^{\text{LR}}_t \oplus S_s(\text{WP}(I^{\text{est}}_{t-1}, \text{UP}(\text{FNet}(I^{\text{LR}}_{t-1}, I^{\text{LR}}_t))))) \quad (6)$$

### 3.2. Loss functions

We use two loss terms to train our model, see Fig. 2, right. The loss $\mathcal{L}_{\text{sr}}$ is applied on the output of SRNet and is backpropagated through both SRNet and FNet:

$$\mathcal{L}_{\text{sr}} = ||I^{\text{est}}_t - I^{\text{HR}}_t||^2_2 \quad (7)$$

Since we do not have a ground truth optical flow for our video dataset, we calculate the spatial mean squared error on the warped LR input frames leading to the auxiliary loss term $\mathcal{L}_{\text{flow}}$ to aid FNet during training.

$$\mathcal{L}_{\text{flow}} = ||\text{WP}(I^{\text{LR}}_{t-1}, F^{\text{LR}}) - I^{\text{LR}}_t||^2_2 \quad (8)$$

The total loss used for training is $\mathcal{L} = \mathcal{L}_{\text{sr}} + \mathcal{L}_{\text{flow}}$.

### 3.3. Justifications

The proposed FRVSR framework is motivated by the following ideas:

- Processing the input video frames more than once leads to high computational cost. Hence, we avoid the sliding window approach and process each input frame only once.

- Having direct access to the previous output can help the network to produce a temporally consistent estimate for the following frame. Furthermore, through a recurrent architecture, the network can effectively use a large number of previous LR frames to estimate the HR frame (see Sec. 4.6) without tradeoffs in computational efficiency. For this reason, we warp the previous HR estimate and feed it to the super-resolution network.

- All computationally intensive operations should be performed in LR space. To this end, we map the previous HR estimate to LR space using the space-to-depth transformation, the inverse of which has been previously used by Shi *et al.* [36] for upsampling. Running SRNet in LR space has the additional advantages of reducing the memory footprint and increasing the receptive field when compared to a super-resolution network that would operate in HR space.

### 3.4. Implementation

The proposed model in Fig. 2 is a flexible framework that leaves the choice for a specific network architecture open. For our experiments, we use fully convolutional architectures for both FNet and SRNet, see Fig. 4 for details. The design of our optical flow network FNet follows a simple encoder-decoder style architecture to increase the receptive field of the convolutions. For SRNet, we follow the residual architecture used by Sajjadi *et al.* [35], but replace the upsampling layers with transposed convolutions. Our choice of network architectures strikes a balance between quality and complexity. More recent methods for each subtask, especially more complex optical flow estimation methods [7, 18, 33] can be easily incorporated and will lead to even better results.

### 3.5. Training and Inference

Our training dataset consists of 40 high-resolution videos (720p, 1080p and 4k) downloaded from vimeo.com. We downsample the original videos by a factor of 2 to have a clean high-resolution ground truth and extract patches of size 256×256 to generate the HR videos. To produce the input LR videos, we apply Gaussian blur to the HR frames and downscale them by sampling every 4-th pixel in each dimension for $s = 4$. Unless specified otherwise, we use a Gaussian blur with standard deviation $\sigma = 1.5$ (see Sec. 4.2).

To train the recurrent system, we extract clips of 10 consecutive frames from the videos using FFmpeg. We avoid cuts or large scene changes in the clips by making sure that the clips do not contain keyframes. All losses are backpropagated through both networks SRNet and FNet as well as through time, *i.e.*, even the optical flow network for the first frame in a clip receives gradients from the super-resolution loss on the 10th frame. The model directly estimates the full RGB video frames, so no post-processing is necessary.

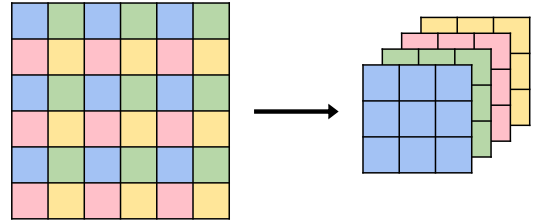

Figure 3: Illustration of the space-to-depth transformation $S_2$. Regular LR grids with varying offsets are extracted from an HR image and placed into the channel dimension, see Eq. 5 for a formal definition.

To estimate the first frame $I_1^{est}$ in each clip, we initialize the previous estimate with a black image $I_0^{est} = 0$ at both training and testing time. The network will then simply upsample the input frame $I_1^{LR}$ independently without additional prior data, similar to a single image super-resolution network. This has the additional benefit of encouraging the network to learn how to upsample single images independently early on during training instead of only relying on copying the previously generated image $\tilde{I}_{t-1}^{est}$.

Our architecture is fully end-to-end trainable and does not require component-wise pre-training. Initializing the networks with the Xavier method [14], we train the model on 2 million batches of size 4 using the Adam optimizer [24] with a fixed learning rate of $10^{-4}$. Note that each sample in the batch is a set of 10 consecutive video frames, *i.e.*, 40 video frames are passed through the networks in each iteration.

As training progresses, the optical flow estimation gradually improves which gives the super-resolution network higher-quality data to work with, helping it to rely more and more on the warped previous estimate $\tilde{I}_{t-1}^{est}$. At the same time, the super-resolution network automatically learns to ignore the previous image $\tilde{I}_{t-1}^{est}$ when the optical flow network cannot find a good correspondence between $I_{t-1}^{LR}$ and $I_t^{LR}$, *e.g.*, for the very first video frame in each batch or for occluded areas. These cases can be detected by the network through a comparison of the low frequencies in $\tilde{I}_{t-1}^{est}$ with those in $I_t^{LR}$. In areas where they do not match, the network ignores the details in $\tilde{I}_{t-1}^{est}$ and simply upscales the current input frame independently. Once the model has been trained, it can be run on videos of arbitrary size and length due to the fully convolutional nature of the networks. To super-resolve a video, the network is applied frame by frame in a single feed-forward pass. Benchmarks for runtimes of different model sizes are reported in Sec. 4.7.

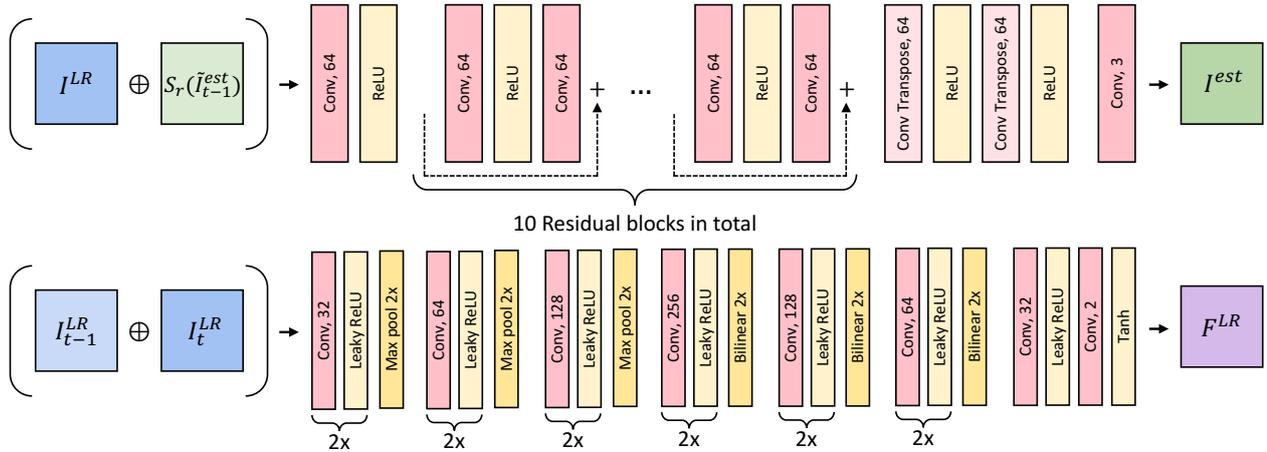

Figure 4: Network architectures for SRNet (top) and FNet (bottom) for 4x upsampling. Both networks are fully convolutional and work in LR space. For the inputs, ⊕ denotes the concatenation of images in the channel dimension. All convolutions in both networks use 3×3 kernels with stride 1, except for the transposed convolutions in SRNet which use stride 2 for spatial upsampling. The leaky ReLU units in FNet use a leakage factor of 0.2 and the notation 2x indicates that the corresponding block is duplicated.

## 4. Evaluation

For a fair evaluation of the proposed framework on equal ground, we compare our model with two baselines that use the same optical flow and super-resolution networks. After presenting the baselines in Sec. 4.1, we extensively investigate the performance of FRVSR along with the baselines in Sec. 4.2–4.7. All experiments are done for the challenging case of 4x upsampling. For evaluation, we use a dataset of ten 3–5s high-quality 1080p video clips downloaded from youtube.com, which we refer to as YT10. Finally, we compare our models with current state-of-the-art methods on the standard Vid4 benchmark dataset [27] in Sec. 4.8. Following Caballero *et al.* [3], we compute *video* PSNR on the brightness channel (ITU-R BT.601 YCbCr standard) using the mean squared error over all pixels in the video.

For more results and video samples, we refer the reader to our homepage at msajjadi.com.

### 4.1. Baselines

**SISR:** For the single image super-resolution baseline, we omit optical flow estimation from FRVSR and disregard any prior information, feeding only $I_t^{\text{LR}}$ into SRNet.

**VSR:** To compare with the sliding window approach for video super-resolution, we include this baseline in which a fixed number of input frames are processed to produce a single output frame. Following Kappeler *et al.* [20] and Caballero *et al.* [3], we warp the previous and next input frames onto the current frame, concatenate all three frames and feed them to SRNet. Note that this model is computationally more expensive than FRVSR since it runs FNet twice for each frame while the computation for SRNet is almost identical to that of FRVSR.

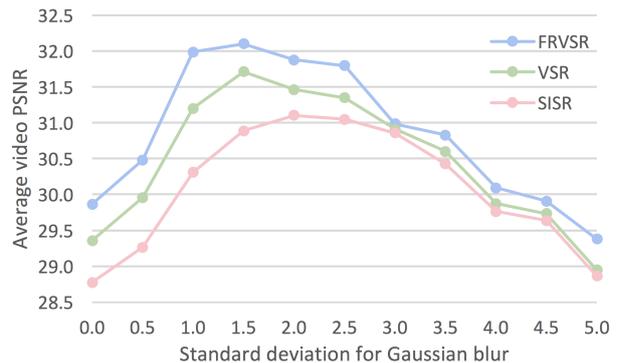

Figure 5: Performance for different blur sizes on YT10. For all blur sizes, FRVSR gives the best results. The best PSNR of FRVSR ($\sigma = 1.5$) is 1.00 dB and 0.39 dB higher than the best of SISR ($\sigma = 2.0$) and VSR ($\sigma = 1.5$), respectively.

As with FRVSR, both baselines are trained starting from a Xavier initialization [14] using the Adam optimizer [24] with a fixed learning rate of $10^{-4}$. We trained the SISR network for 500K steps and VSR for 2 million steps, both using a batch size of 16. All networks are trained using the same dataset, and their losses on a validation dataset have converged at the end of the training.

### 4.2. Blur size

As mentioned in Sec. 3.5, we apply Gaussian blur to the HR frames before downsampling them to generate the LR input for the network. While a smaller blur kernel results in aliasing, excessive blur leads to loss of high-frequency information in the input, making it harder to reconstruct finer details. To analyze how different approaches perform for blurry or aliased inputs, we trained SISR, VSR and FRVSR on video frames that have been downscaled using different values of standard deviation for the Gaussian blur ranging from $\sigma = 0$ to $\sigma = 5$, see Fig. 5. The proposed framework FRVSR significantly outperforms SISR and VSR on all blur sizes. It is interesting to note that SISR, which relies on a single LR image for upsampling, benefits the most from larger blur kernels compared to VSR and FRVSR which perform best with $\sigma = 1.5$. This is due to the fact that video super-resolution methods are able to blend information from multiple frames and therefore benefit from sharper inputs. In the remaining experiments, we use a value of $\sigma = 1.5$.

### 4.3. Training clip length

Since FRVSR is a recurrent network, it can be trained on video clips of any length. To test the effect of the clip length used to train the network on its performance, we trained the same model using video clips of length 2, 5 and 10, yielding average video PSNR values of 31.60, 32.01 and 32.10 on YT10, respectively. These results show that the PSNR has already started to saturate with a clip length of 5 and going beyond 10 may not yield significant improvements.

### 4.4. Degraded inputs

To see how different models perform under input degradations, we trained and evaluated FRVSR and the baselines using noisy and compressed input frames. Table 1 shows the performance of these models on YT10 for varying levels of Gaussian noise and JPEG compression quality. The proposed framework consistently outperforms both SISR and VSR by 0.36–0.91 dB and 0.18–0.48 dB, respectively.

### 4.5. Temporal consistency

Analyzing the temporal consistency of the results is best done by visual inspection of the video results. However, to compare the results on paper, we follow Caballero *et al.* [3] and show *temporal profiles*, see Fig. 6. A temporal profile is generated by taking the same horizontal row of pixels from a number of frames in the video and stacking them vertically into a new image. Flickering in the video will show up as jitter and jagged lines in the temporal profile. While VSR produces sharper results than SISR, it still has significant flickering artifacts since each output frame is estimated separately. In contrast, FRVSR produces the most consistent results while containing even finer details in each image.

### 4.6. Range of information flow

Existing approaches to video super-resolution often use a fixed number (usually 3–7) of input frames to produce a single output frame. Increasing this number increases the maximum number of frames over which details can be propagated. While this can result in higher-quality videos, it also substantially increases the computational cost, leading to a tradeoff between efficiency and quality. In contrast, due to its recurrent nature, FRVSR can pass information across a large number of frames without increasing computations. Figure 7 shows the performance of FRVSR as a function of the number of frames processed. In the normal mode (blue curve) in which a black frame is used as the first frame's previous HR estimate, the performance steadily improves as more frames are processed and it plateaus at 12 frames. When we replace the first previous HR estimate with the corresponding groundtruth HR frame (red curve), FRVSR carries the high-frequency details across a large number of frames and performs better than the normal mode even after 50 frames.

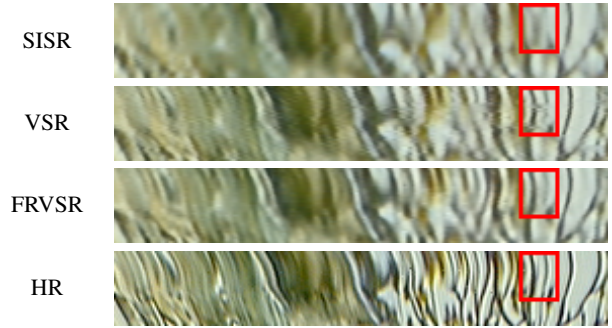

Figure 6: Temporal profiles for *Calendar* from Vid4. VSR yields finer details than SISR, but it's output still contains temporal inconsistencies (see red boxes). Only FRVSR is able to produce temporally consistent results while reproducing fine details. Best viewed on screen.

| model | $\sigma = 0.025$ | $\sigma = 0.075$ | JPG 40 | JPG 70 |
|---|---|---|---|---|
| SISR | 29.93 | 28.20 | 27.94 | 28.88 |
| VSR | 30.36 | 28.42 | 28.12 | 29.07 |
| FRVSR | **30.84** | **28.62** | **28.30** | **29.29** |

Table 1: Average video PSNR of various models under Gaussian noise (left) and JPEG artifacts (right) on YT10. In all experiments, FRVSR achieves the highest PSNR.

To investigate the maximum effective range of information flow, we start the same model at different input frames in the same video and compare the performance. Figure 8 shows such a comparison for the *Foliage* video from Vid4. As we can see, the gap between the curves for the models that start at frame 1 and frame 11 only closes towards the end of the clip, showing that FRVSR is propagating information over more than 30 frames. To propagate details over such a large range, previous state-of-the-art methods [3, 20, 28, 29, 39] would have to process an inhibiting number of input frames for each output image, which would be computationally infeasible.

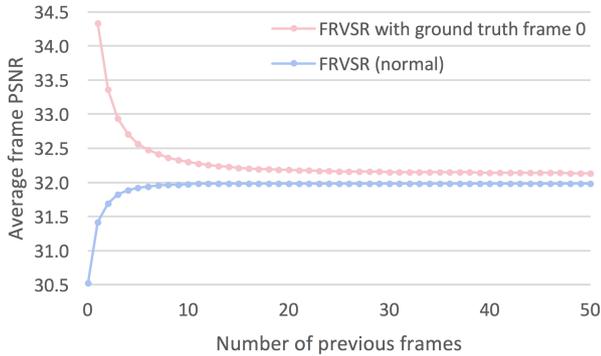

Figure 7: Performance of FRVSR on YT10 as a function of the number of previous frames processed. In the normal mode (blue), PSNR increases up to 12 frames, after which it remains stable. When the first HR image is given (red), FRVSR propagates high-frequency details across a large number of frames and performs better than the normal mode even after 50 frames.

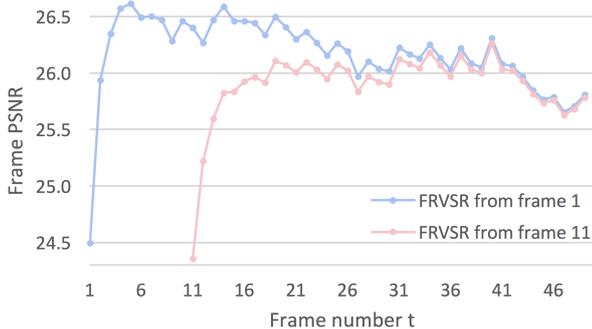

Figure 8: Performance of FRVSR started at the 1st and 11th frame of *Foliage* from Vid4. The gap between the curves only closes towards the end of the clip, showing FRVSR's ability to retain details over a large range of video frames.

### 4.7. Network size and computational efficiency

To see how the performance of different models varies with the size of the network, we trained and evaluated FRVSR and the baselines with different numbers of residual blocks and convolution filters in SRNet, see Fig. 9. It is interesting to note that the video super-resolution models FRVSR and VSR clearly benefit from larger models while the performance of SISR does not change significantly beyond 5 residual blocks. We can also see that FRVSR achieves better results than VSR despite being faster: The FRVSR models with 5 residual blocks outperform the VSR models with 10 residual blocks, and the FRVSR models with 3 residual blocks outperform the VSR models with 5 residual blocks for the same number of convolution filters.

With our unoptimized TensorFlow implementation on an Nvidia P100, producing a single Full HD frame for 4x up-scaling takes 74ms for FRVSR with 3 residual blocks and 64 filters, and 191ms for FRVSR with 10 blocks and 128 filters.

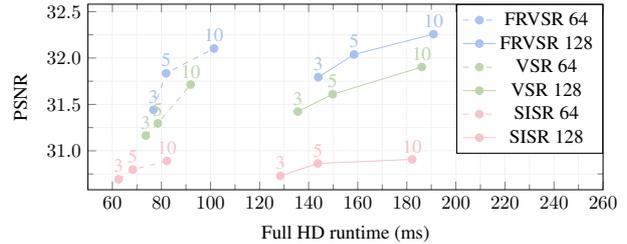

Figure 9: Performance on YT10 for different numbers of convolution filters (64 / 128) and residual blocks in SRNet. FRVSR achieves better results than both baselines with significantly smaller super-resolution networks and less computation time. For example, FRVSR with 5 residual blocks is both faster and better than VSR with 10 residual blocks.

### 4.8. Comparison with prior art

Table 2 compares the proposed FRVSR approach with various state-of-the-art video super-resolution approaches on the standard Vid4 benchmark dataset by PSNR and SSIM. We report results for two FRVSR networks: FRVSR 10-128, which is our best model with 10 residual blocks and 128 convolution filters, and FRVSR 3-64, which is our most efficient model with only 3 residual blocks and 64 convolution filters. For the baselines SISR and VSR, we report their best results which correspond to 10 residual blocks and 128 convolution filters. We also include RAISR [34] as an off-the-shelf single image super-resolution alternative.

For all competing methods except [3, 17, 34], we used the output images provided by the corresponding authors to compute PSNR and SSIM. We did not use the first and last two frames in our evaluation since Liu *et al.* [28] do not produce outputs for these frames. Also, for each video, we removed border regions such that the LR input image is a multiple of 8. For [3, 17], we use the PSNR and SSIM values reported in the respective publications since we could not confirm them independently. For [34], we used the models provided by the authors to generate the output images.

As shown in Tab. 2, FRVSR outperforms the current state of the art by more than 0.5 dB. In fact, even our most efficient model FRVSR 3-64 produces state-of-the-art results by PSNR and beats all previous neural network-based methods by SSIM. It it interesting that our small model, despite being much more efficient, produces results that are very close to the much larger model VSR 10-128 on the Vid4 dataset.

Figure 10 shows a visual comparison of the different approaches. We can see that our models are able to recover fine details and produce visually pleasing results. Even our most efficient network FRVSR 3-64 produces higher-quality results than prior art.

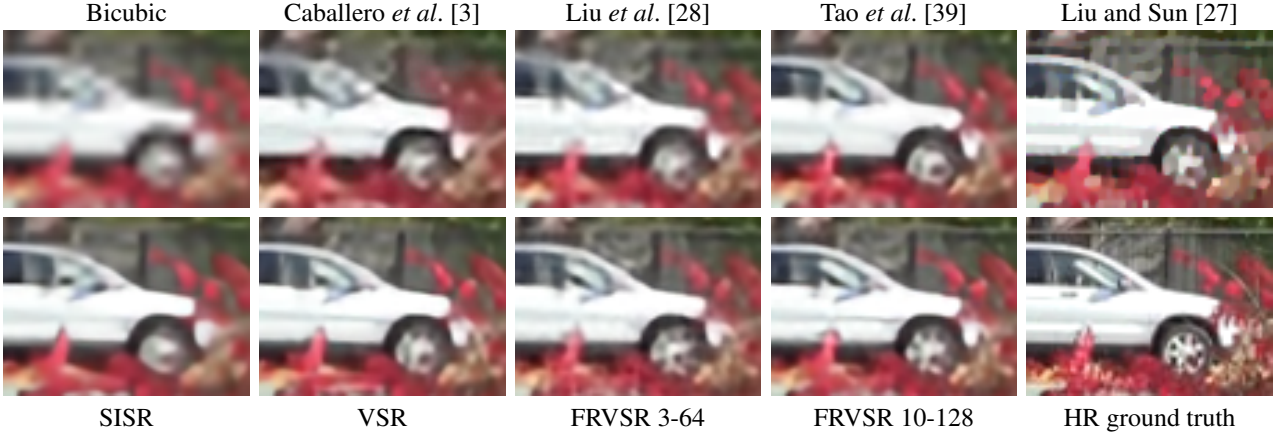

Figure 10: Visual comparison with previous methods on *Foliage* from Vid4. Amongst prior art, Liu and Sun [27] recover the finest details, but their result has blocky artifacts, and their method uses a slow optimization procedure. Between the remaining methods, even the result of our smallest model FRVSR 3-64 is sharper and contains more details than prior art, producing results similar to the much bigger VSR model. Our larger model FRVSR 10-128 recovers the most accurate image.

| Method | Bicubic | RAISR [34] | BRCN [17] | VESPCN [3] | $B_{1,2,3}+T$ [28] | DRVSR [39] | Bayesian [27] | SISR 10-128 | VSR 10-128 | FRVSR 3-64 | FRVSR 10-128 |
|---|---|---|---|---|---|---|---|---|---|---|---|
| PSNR | 23.53 | 24.24 | 24.43* | 25.35* | 25.35 | 25.87 | 26.16 | 24.96 | 26.25 | 26.17 | **26.69** |
| SSIM | 0.628 | 0.665 | 0.662* | 0.756* | 0.738 | 0.772 | 0.815 | 0.721 | 0.803 | 0.798 | **0.822** |

Table 2: Comparison of average PSNR and SSIM on the standard Vid4 dataset for scaling factor $s=4$. Our smallest model FRVSR 3-64 already produces better results than all prior art including the computationally expensive optimization-based method by Liu and Sun [27] by PSNR. Using a bigger super-resolution network helps FRVSR 10-128 to add an additional 0.5 dB on top and achieve state-of-the-art results by SSIM as well, showing that the proposed framework can greatly benefit from more powerful networks. Values marked with a star have been copied from the respective publications.

## 5. Future work

Since our framework relies on the HR estimate $I^{\text{est}}$ for propagating information, it can reconstruct details and propagate them over a large number of frames (see Sec. 4.6). At the same time, any detail can only persist in the system as long as it is contained in $I^{\text{est}}$, as it is the only way through which SRNet can pass information to future iterations. Due to the spatial loss on $I^{\text{est}}$, SRNet has no way to pass on auxiliary information that could potentially be useful for future frames in the video, *e.g.*, for occluded regions. As a result, occlusions irreversibly destroy all previously aggregated details in the affected areas and the best our model can do for the previously occluded areas is to match the performance of single image super-resolution models. In contrast, models that use a fixed number of input frames can still combine information from frames that do not have occlusions to produce better results in these areas. To address this limitation, it is natural to extend the framework with an additional memory channel. However, preliminary experiments in this direction with both static and motion-compensated memory did not improve the overall performance of the architecture, so we leave further investigations in this direction to future work.

Since the model is conceptually flexible, it can be easily extended to other applications. As an example, one may plug in the original HR frame $I^{\text{HR}}_{t-1}$ in place of the estimated frame $I^{\text{est}}_{t-1}$ for every $K$-th frame. This could enable an efficient video compression method where only one in $K$ HR-frames needs to be stored while the remaining frames would be reconstructed by the model.

A further extension of our framework would be the inclusion of more advanced loss terms which have recently been shown to produce more visually pleasing results [26, 35]. The recurrent architecture in FRVSR naturally encourages the network to produce temporally consistent results, making it an ideal candidate for further research in this direction.

## 6. Conclusion

We propose a flexible end-to-end trainable framework for video super-resolution that is able to generate higher quality results while being more efficient than existing sliding window approaches. In an extensive set of experiments, we show that our model outperforms competing baselines in various different settings. The proposed model also significantly outperforms state-of-the-art video super-resolution approaches both quantitatively and qualitatively on a standard benchmark dataset.

# References

[1] E. Agustsson and R. Timofte. NTIRE 2017 challenge on single image super-resolution: Dataset and study. In *CVPR workshops*, 2017.

[2] S. P. Belekos, N. P. Galatsanos, and A. K. Katsaggelos. Maximum a posteriori video super-resolution using a new multi-channel image prior. *IEEE Transactions on Image Processing*, 2010.

[3] J. Caballero, C. Ledig, A. Aitken, A. Acosta, and Totz. Real-time video super-resolution with spatio-temporal networks and motion compensation. In *CVPR*, 2017.

[4] D. Chen, J. Liao, L. Yuan, N. Yu, and G. Hua. Coherent online video style transfer. In *ICCV*, 2017.

[5] C. Dong, C. C. Loy, K. He, and X. Tang. Learning a deep convolutional network for image super-resolution. In *ECCV*, 2014.

[6] C. Dong, C. C. Loy, and X. Tang. Accelerating the super-resolution convolutional neural network. In *ECCV*, 2016.

[7] A. Dosovitskiy, P. Fischer, E. Ilg, P. Hausser, C. Hazirbas, V. Golkov, P. van der Smagt, D. Cremers, and T. Brox. Flownet: Learning optical flow with convolutional networks. In *ICCV*, 2015.

[8] M. Drulea and S. Nedevschi. Total variation regularization of local-global optical flow. In *ITSC*, 2011.

[9] C. E. Duchon. Lanczos filtering in one and two dimensions. *Journal of Applied Meteorology*, 1979.

[10] S. Farsiu, M. Elad, and P. Milanfar. Video-to-video dynamic super-resolution for grayscale and color sequences. *EURASIP Journal on Applied Signal Processing*, 2006.

[11] S. Farsiu, M. D. Robinson, M. Elad, and P. Milanfar. Fast and robust multiframe super resolution. *IEEE Transactions on Image Processing*, 2004.

[12] G. Freedman and R. Fattal. Image and video upscaling from local self-examples. *ACM Transactions on Graphics*, 2011.

[13] W. T. Freeman, T. R. Jones, and E. C. Pasztor. Example-based super-resolution. *IEEE Computer Graphics and Applications*, 2002.

[14] X. Glorot and Y. Bengio. Understanding the difficulty of training deep feedforward neural networks. In *AISTATS*, 2010.

[15] A. Gupta, J. Johnson, A. Alahi, and L. Fei-Fei. Characterizing and improving stability in neural style transfer. In *ICCV*, 2017.

[16] J.-B. Huang, A. Singh, and N. Ahuja. Single image super-resolution from transformed self-exemplars. In *CVPR*, 2015.

[17] Y. Huang, W. Wang, and L. Wang. Bidirectional recurrent convolutional networks for multi-frame super-resolution. In *NIPS*, 2015.

[18] E. Ilg, N. Mayer, T. Saikia, M. Keuper, A. Dosovitskiy, and T. Brox. Flownet 2.0: Evolution of optical flow estimation with deep networks. *CVPR*, 2017.

[19] M. Jaderberg, K. Simonyan, A. Zisserman, and K. Kavukcuoglu. Spatial transformer networks. In *NIPS*, 2015.

[20] A. Kappeler, S. Yoo, Q. Dai, and A. K. Katsaggelos. Video super-resolution with convolutional neural networks. In *IEEE Transactions on Computational Imaging*, 2016.

[21] J. Kim, J. Kwon Lee, and K. Mu Lee. Accurate image super-resolution using very deep convolutional networks. In *CVPR*, 2016.

[22] J. Kim, J. Kwon Lee, and K. Mu Lee. Deeply-recursive convolutional network for image super-resolution. In *CVPR*, 2016.

[23] T. H. Kim, K. M. Lee, B. Schölkopf, and M. Hirsch. Online video deblurring via dynamic temporal blending network. In *ICCV*, 2017.

[24] D. Kingma and J. Ba. Adam: A method for stochastic optimization. In *ICLR*, 2015.

[25] W.-S. Lai, J.-B. Huang, N. Ahuja, and M.-H. Yang. Deep laplacian pyramid networks for fast and accurate super-resolution. In *CVPR*, 2017.

[26] C. Ledig, L. Theis, F. Huszár, J. Caballero, A. Aitken, A. Tejani, J. Totz, Z. Wang, and W. Shi. Photo-realistic single image super-resolution using a generative adversarial network. *CVPR*, 2017.

[27] C. Liu and D. Sun. A bayesian approach to adaptive video super resolution. In *CVPR*, 2011.

[28] D. Liu, Z. Wang, Y. Fan, X. Liu, Z. Wang, S. Chang, and T. Huang. Robust video super-resolution with learned temporal dynamics. In *CVPR*, 2017.

[29] O. Makansi, E. Ilg, and T. Brox. End-to-end learning of video super-resolution with motion compensation. In *GCPR*, 2017.

[30] P. Milanfar. *Super-resolution Imaging*. CRC press, 2010.

[31] K. Nasrollahi and T. B. Moeslund. Super-resolution: A comprehensive survey. *Machine Vision and Applications*, 2014.

[32] E. Perez-Pellitero, J. Salvador, J. Ruiz-Hidalgo, and B. Rosenhahn. PSyCo: Manifold span reduction for super resolution. In *CVPR*, 2016.

[33] A. Ranjan and M. J. Black. Optical flow estimation using a spatial pyramid network. *CVPR*, 2017.

[34] Y. Romano, J. Isidoro, and P. Milanfar. RAISR: Rapid and accurate image super resolution. *IEEE Transactions on Computational Imaging*, 2016.

[35] M. S. M. Sajjadi, B. Schölkopf, and M. Hirsch. EnhanceNet: Single image super-resolution through automated texture synthesis. In *ICCV*, 2017.

[36] W. Shi, J. Caballero, F. Huszár, J. Totz, A. P. Aitken, R. Bishop, D. Rueckert, and Z. Wang. Real-time single image and video super-resolution using an efficient sub-pixel convolutional neural network. In *CVPR*, 2016.

[37] Y. Tai, J. Yang, and X. Liu. Image super-resolution via deep recursive residual network. In *CVPR*, 2017.

[38] H. Takeda, P. Milanfar, M. Protter, and M. Elad. Super-resolution without explicit subpixel motion estimation. *IEEE Transactions on Image Processing*, 2009.

[39] X. Tao, H. Gao, R. Liao, J. Wang, and J. Jia. Detail-revealing deep video super-resolution. In *ICCV*, 2017.

[40] R. Timofte, R. Rothe, and L. Van Gool. Seven ways to improve example-based single image super resolution. In *CVPR*, 2016.

[41] C.-Y. Yang, J.-B. Huang, and M.-H. Yang. Exploiting self-similarities for single frame super-resolution. In *ACCV*, 2010.

[42] J. Yang, Z. Lin, and S. Cohen. Fast image super-resolution based on in-place example regression. In *CVPR*, 2013.

[43] J. Yang, Z. Wang, Z. Lin, S. Cohen, and T. Huang. Coupled dictionary training for image super-resolution. *IEEE Transactions on Image Processing*, 2012.